# Vision-based Tactile Image Generation via Contact Condition-guided Diffusion Model

Xi Lin, Weiliang Xu, Yixian Mao, Jing Wang, Meixuan Lv, Lu Liu, Xihui Luo, Xinming Li*

*Abstract*—Vision-based tactile sensors, through high-resolution optical measurements, can effectively perceive the geometric shape of objects and the force information during the contact process, thus helping robots acquire higher-dimensional tactile data. Vision-based tactile sensor simulation supports the acquisition and understanding of tactile information without physical sensors by accurately capturing and analyzing contact behavior and physical properties. However, the complexity of contact dynamics and lighting modeling limits the accurate reproduction of real sensor responses in simulations, making it difficult to meet the needs of different sensor setups and affecting the reliability and effectiveness of strategy transfer to practical applications. In this letter, we propose a contact-condition guided diffusion model that maps RGB images of objects and contact force data to high-fidelity, detail-rich vision-based tactile sensor images. Evaluations show that the three-channel tactile images generated by this method achieve a 60.58% reduction in mean squared error and a 38.1% reduction in marker displacement error compared to existing approaches based on lighting model and mechanical model, validating the effectiveness of our approach. The method is successfully applied to various types of tactile vision sensors and can effectively generate corresponding tactile images under complex loads. Additionally, it demonstrates outstanding reconstruction of fine texture features of objects in a Montessori tactile board texture generation task.

## I. INTRODUCTION

Accurately capturing and processing real tactile signals is crucial in the interaction between robots and the external physical world [1]. In recent years, various types of vision-based tactile sensors have been widely applied in robotic tasks [2]-[4]. Vision-based tactile sensors, through high-resolution optical measurements, can effectively perceive the geometry of objects and force information during the contact process, thereby helping robots acquire higher-dimensional tactile data. For data-driven perception and control tasks with vision-based tactile sensors, a common reinforcement learning approach is to train robots in simulated environments to avoid time-consuming and labor-intensive experiments in real environments, enabling more effective transfer of learned strategies to real robots [5]-[7]. However, the complexity of contact requirements limits the accuracy of the model because accurately simulating such tactile sensors not only requires reproducing the dynamic characteristics of the contact process but also necessitates the development of the sensor's lighting model to ensure realistic perceptual output. This leads to simulations that cannot accurately represent the optical and mechanical responses in complex contact scenarios, further affecting the reliability and effectiveness of the trained strategies in real-world applications. At the same time, the additional model design requirements for different configurations of vision-based tactile sensors further increase the complexity of simulations, making it more challenging to develop a universal training strategy [8]-[11].

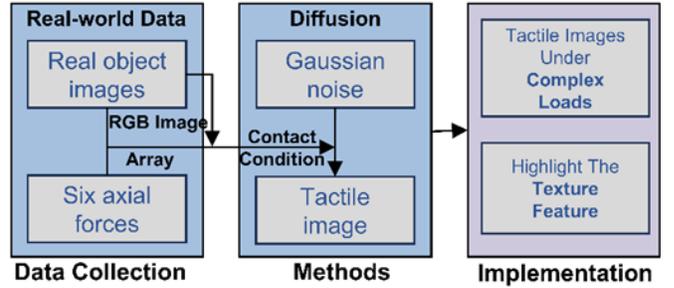

**Fig. 1.** The proposed universal method workflow for multi-structure vision-based tactile sensors

To address the challenges mentioned above, this work adopts a data-driven approach to convert real object images and contact forces into RGB tactile images captured by real vision-based tactile sensors. The principle is based on the actual generation mechanism of vision-based tactile data, which allows us to reveal the generation conditions and logical relationships of tactile images [12]. Specifically, tactile images can directly reflect the 3D shape and texture of an object's surface, and by analyzing the deformation areas, this approach can infer key information such as the physical properties of the contacted object and the detected contact forces [13]-[17]. Our research approaches the problem from the perspective of data generation, using real data to reverse-engineer the process of tactile image generation. By analyzing the generation conditions of tactile images, we can reverse-engineer the corresponding vision-based tactile sensor images, thus eliminating the need for optical and mechanical modeling.

This work was supported by the Guangdong Basic and Applied Basic Research Foundation (No. 2022A1515010136), the Guangdong Provincial Key Laboratory of Nanophotonic Functional Materials and Devices, and the South China Normal University start-up fund. *(Xi Lin and Weiliang Xu contributed equally, Corresponding author: Xinming Li).*

Xi Lin, Weiliang Xu, Yixian Mao, Jing Wang, Meixuan Lv, Lu Liu, Xihui Luo and Xinming Li are with Guangdong Provincial Key Laboratory of Nanophotonic Functional Materials and Devices, Guangdong Basic Research Center of Excellence for Structure and Fundamental Interactions of Matter, School of Optoelectronic Science and Engineering, South China Normal University, Guangzhou 510006, China. (e-mail: xmli@m.scnu.edu.cn)

To address the challenge of accurately generating tactile images that capture intricate surface textures and contact behaviors, a condition-guided diffusion model is adopted. This approach leverages its ability to iteratively refine image quality through a noise redistribution process, guided by contact conditions. Thus, this letter, based on a condition-guided diffusion model, generates tactile images from real data using contact information. It employs real object RGB images and six-axis force data arrays as conditions to guide the redistribution of Gaussian noise pixel sequences, generating tactile images that correspond to contact behaviors. This approach generates high-fidelity tactile images starting from real data and restores details such as the surface texture of objects (Fig. 1). In addition, this method can generate vision-based tactile sensor images with varying lighting conditions under different force scenarios, effectively narrowing the gap between simulation and reality and fostering advancements in tactile-motor robotic manipulation skills. In summary, the contributions of this work are three-fold:

- A new contact condition-guided diffusion model approach is introduced for pixel-level data mapping between different data domains, learning the sensor's optical environment and the elastomer's deformation motion. The RGB-lighting tactile images generated by this method exhibit a mean squared error (MSE) that is 62.97% lower compared to existing approaches based on lighting model and mechanical model [18], [19].

- Our method is applied to various custom vision-based tactile sensors, such as photometric stereo and marker-based systems. Under different contact loads, this approach outperforms other tactile image simulators in terms of marker displacement error metrics, achieving a 55.61% reduction in marker displacement error compared to existing approaches based on lighting model and mechanical model [18], [20].

- We applied this model to the texture generation task of Montessori tactile boards. The model's high-fidelity capability enhances the generation of fine object details, demonstrating its effectiveness in accurately restoring subtle texture features.

## II. RELATED WORK

### A. Simulation of Vision-Based Tactile Sensor

Vision-based tactile sensors, in addition to high-resolution measurements of the geometric shapes of contact objects, can acquire more dimensional physical properties and contact-related physical information by altering the sensor structure [12], [21], [22]. Various vision-based tactile sensors have been developed based on different operating principles to capture higher-dimensional tactile information. For example, multi-color light sensors based on photometric stereo [23], [24] algorithms estimate the surface normal of deformed areas under different lighting conditions to infer the three-dimensional shape of the contact object; vision-based tactile sensors with markers deduce the magnitude and direction of contact forces by calculating the displacement of the markers [25]. To address the limitations of real sensors in data-driven methods, simulation allows for the accurate capture and analysis of contact behaviors and physical properties without the need for physical sensors, thereby effectively supporting the acquisition and understanding of tactile information. Current research on vision-based tactile sensor simulations continues to advance, with simulation playing a critical role in emulating and optimizing the tactile information acquisition process without relying on physical sensors. Some studies utilize simulated depth images to reconstruct the interactions between contact objects and elastic materials, along with optical rendering [26], [27]. However, due to the Sim2Real gap caused by the lack of real-world feature modeling in simulation, using simulated depth images to replace elastomer representations results in insufficient detail reproduction when simulating real-world physical properties and fails to capture complex force feedback accurately. Tacto [28] and Taxim [18] have introduced elastomer physical models. However, since they rely on finite element methods [29], [30] or specific physical models, adjustments to the light field and elastomer model parameters are required for different sensors, and they lack generalization capability for various contact objects (flat and non-flat).

Our method relies on real tactile sensor data, preserving the inherent noise within the sensor to narrow the gap between reality and simulation, rather than analyzing physical contact models or training generative networks. This letter introduces an effective method that infers tactile images from contact conditions by transforming contact condition information into corresponding vision-based tactile sensor images of different types, accurately approximating elastomer motion under various load types.

### B. Diffusion Model

A diffusion model is a parameterized Markov chain trained using variational inference, which learns to transform a standard normal distribution into an empirical data distribution, thereby generating samples matching the data after a finite time [31], [32]. In the field of vision-based tactile sensor, experiments aim to generate vision-based tactile sensor images of various types under different loads based on contact conditions, achieving closer alignment between simulation and reality. This requires implementation through transformations across data domains. By leveraging the generative capability of diffusion models and cross-domain data mapping, we can generate vision-based tactile sensor images under specific conditions, capturing and reconstructing the response characteristics of light and elastic force fields under varying loads.

Our approach directly utilizes real data, conditioning the diffusion model on pairs of contact information comprising real object images and contact forces. This guides the generation of RGB tactile images under varying loads, faithfully reconstructing the texture features from the real object images.

## III. METHODS

The proposed method develops a general generation method for vision-based tactile sensors with various types under complex loads through data collection, training, and accuracy and quality of the generated images. Additionally, experiments implementation. First, we collect object images and their corresponding tactile images under specific loads. The content of the contact conditions is explained in Section III-A. Subsequently, we use the dataset to train the diffusion model, ultimately achieving the direct generation of high-fidelity tactile images from contact information.

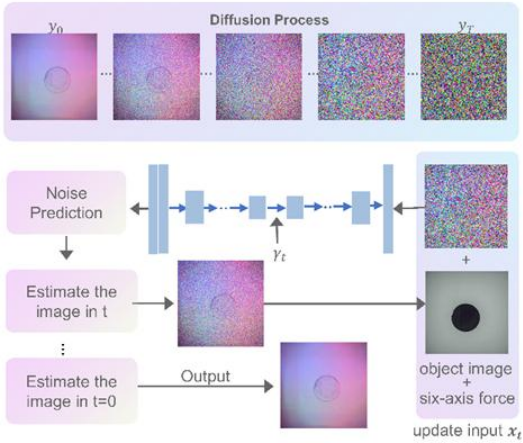

**Fig. 2.** The pipeline of training contact condition-guided diffusion model

*A. Contact Conditions*

Our approach does not require optical or mechanical modeling by directly mapping contact conditions to tactile images. When an object contacts the sensor, the soft elastomer layer deforms, resulting in changes in light emission from the reflective layer on the elastomer surface. The CMOS camera inside the sensor captures these changes. By analyzing the distribution patterns of the emission field intensity in the high-resolution tactile images collected, the geometry and contact force of the contacting object can be decoupled.

The shape and textural features of the shadow regions in tactile images depend on the posture, position, and surface texture of the contacting object. Starting from real data, the geometric shape, posture, position, and surface texture of the object can be intuitively obtained by analyzing the real images of the object. These features are readily apparent in the images and can be recognized without complex processing. Additionally, the brightness distribution and edge contrast of the shadow regions in tactile images are determined by the direction and magnitude of the contact forces. These forces can be decomposed into multi-axis forces, a method that not only captures the complexity of the contact behavior but also provides a more rigorous description of the contact process. Moreover, the contact force information can be intuitively added using mathematical models. Thus, our approach uses pairs of real images of the contacting object and six-axis force arrays as data, guiding the model to generate corresponding tactile images based on real contact conditions. This reduces the gap between generated tactile images and real tactile images. By using a data-driven approach, we avoid the need to re-adjust optical and mechanical models for different types of vision-based tactile sensor.

*B. Condition-Guided Diffusion Model*

Based on the aforementioned mapping relationship between contact conditions and tactile sensor images; to achieve information conversion across different data domains, we use a conditional-guided diffusion model to learn the pixel-level mapping between contact conditions (including contact object images and six-axis contact forces) and tactile images. Specifically, the overall process of this method is shown in Fig. 2: tactile images first undergo the addition of Gaussian noise through T time steps during the diffusion process, and then, under the guidance of the contact conditions, the noise is iteratively removed while adjusting the pixel distribution, thereby achieving the mapping from contact conditions to the target tactile image. The architecture of this method allows the model to effectively capture the transformation rules across different data domains and precisely generate tactile images that align with specific contact conditions.

The object contact state information is composed of the real image of the object and the contact force concatenated together. The mathematical expression:

$$x = concat\left(I, \ H(F)\right) \quad (1)$$

where $I$ is the real image tensor of the contacting object with dimensions $(3, 256, 256)$, and $F$ is a one-dimensional sequence, which cannot be directly combined with the image tensor $I$. Therefore, we use a hash function $H(\cdot)$ to expand $F$ into a tensor of size $(1, 256, 256)$ before concatenation. The hash function is a mathematical method that maps input data of any length to an output of fixed length, ensuring a consistent size and extracting core features. The model then inputs the tensor of the size $(4, 256, 256)$ into the U-Net architecture, where iterative refinement redistributes the pixel tensor, approximating the tactile image $y_0$ corresponding to the contact conditions. The iterative process of pixel redistribution guided by contact conditions can be expressed as $f_\theta(x, y_t, \gamma_t)$, where, in addition to the contact conditions $x$ and the noise image output $y_t$, the statistical noise quantity $\gamma_t$ is also included. The model's loss function can be expressed as:

$$L(\theta) = \mathbb{E}_{(x,y_0)} \mathbb{E}_{\epsilon,t}[\|\epsilon - f_\theta(x, y_t, \gamma_t)\|^2] \quad (2)$$

For different types of vision-based tactile sensors, the model does not need to be modified; it can be adapted by training on datasets specific to each type. Once trained, the model can generate corresponding tactile images based on the object's contact image and six-axis force data.

## IV. EXPERIMENTS AND RESULTS ANALYSIS

To bridge the gap between simulation and reality, this letter adopts a data-driven approach based on real data. Initially, image data from various types of vision-based tactile sensors, along with their corresponding contact condition information, were collected to provide authentic data for model training. In the validation phase of the generated tactile images, the model's performance in reconstructing contact information is evaluated by calculating the similarity between the generated and real images. For optical tactile sensors with markers, a marker displacement error metric is introduced to further quantify the are designed to showcase the model's capability in generating intricate tactile textures, demonstrated using Montessori tactile perception boards. Finally, a visual comparison of the generated images with those produced by other methods highlights the significant advantages of real-data-based

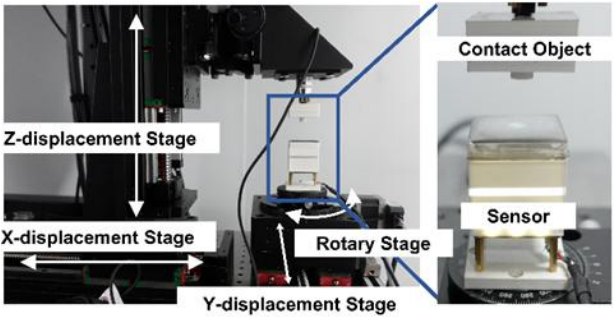

**Fig. 3.** Data acquisition system

generation approaches in simulating real-world physical scenarios.

*A. Data Collection and Training*

To collect the contact images and contact forces corresponding to the tactile images of the object, we constructed a data acquisition system on an optical platform, as shown in Fig. 3

The acquisition system consists of a force gauge, a displacement platform, and sensors. The three displacement stages can apply forces in the x, y, and z directions, while the rotary stage can apply torsion. The force gauge is mounted on the z-axis displacement stage, and the contacting object is fixed to the probe of the force gauge, which can measure the six-axis forces during contact with the sensor $(F_x, F_y, F_z, M_x, M_y, M_z)$, Here, $F_x$, $F_y$, $F_z$, are the forces in the x, y, and z axes, measured in Newtons (N), and $(M_x, M_y, M_z)$ are the torques in the x, y, and z axes, measured in Newton-millimeters (N·mm). We control the displacement stages to make the object contact the sensor, defining the initial contact point as (0 mm, 0 mm, 0 mm). After the object reaches the initial point, it is displaced downward by 0.4 mm, then moved 2 mm along the y-axis to reach (0 mm, 2 mm, -0.4 mm), followed by a 2 mm displacement in the x-axis direction to (0 mm, 2 mm, -0.4 mm), followed by a 2 mm displacement in the x-axis direction to (2 mm, 2 mm, -0.4 mm). After returning to the origin, the same movement strategy is executed in the opposite direction along the x and y axes. Finally, it returns to (0 mm, 0 mm, -0.4mm) and is twisted by 5 degrees around the z-axis. This process is repeated until the z-axis position reaches -1.6 mm. During the movement, tactile images from the vision-based tactile sensor and the corresponding six-axis force output from the force sensor are simultaneously collected at a frequency of 1 Hz by triggering external signals. We executed the movement process slowly (0.01 mm/s), allowing sufficient time between deformations to eliminate rate-dependent nonlinear effects.

This study collected pairs of data comprising vision-based tactile sensor images and contact state information (contact object images and contact forces) for different objects, with image sizes of 3×256×256 and approximately 700 pairs of contact information data for each object. We divided the dataset into training and testing sets in an 8:2 ratio. All training and testing computations were performed using an NVIDIA 3080 GPU, with an average time of 0.26 seconds per training iteration.

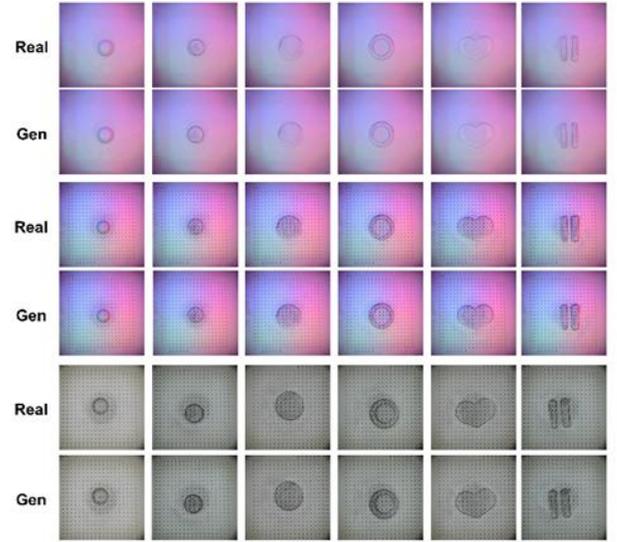

**Fig. 4.** Real images and generation effects of vision-based tactile sensors with different types

TABLE I

THE COMPARISON OF IMAGE SIMILARITY MEASUREMENTS BETWEEN REAL AND GENERATED TACTILE IMAGES OF WITH DIFFERENT TYPES

|  | MSE | MAE | SSIM | PSNR |
| --- | --- | --- | --- | --- |
| TACTO[28](RGB-light) | 201.87 | 10.31 | 0.80 | 25.43 |
| Taxim[18] (RGB-light) | 56.715 | 5.14 | 0.88 | 31.12 |
| Fots[19] (RGB-light) | 53.267 | 4.91 | 0.89 | 32.30 |
| Ours(RGB-light) | 21.00 | 3.39 | 0.97 | 36.62 |
| Ours(RGB-light Mark) | 28.02 | 3.24 | 0.95 | 34.28 |
| Ours(White-light Mark) | 67.09 | 5.27 | 0.81 | 30.91 |

*B. Image Similarity Analysis*

After completing model training, the diffusion model guided by contact conditions was evaluated and compared for its ability to generate tactile images under varying forces. The analysis focused on the similarity between generated images and real sensor images across different types of sensors. Results demonstrate that the proposed method effectively captures the features of contact conditions and efficiently generates image data closely resembling those of real sensors. Fig. 4 illustrates the generated image results for different sensors, while Table 1 provides quantitative comparisons, further validating the method's reliability and applicability in tactile image generation. We used four image similarity metrics to assess and compare the generated images: Mean Absolute Error (MAE), Mean Squared Error (MSE), Structural Similarity Index Measure (SSIM), and Peak Signal-to-NoiseRatio (PSNR). MAE and MSE evaluate the pixel value differences between the generated image and the real image.

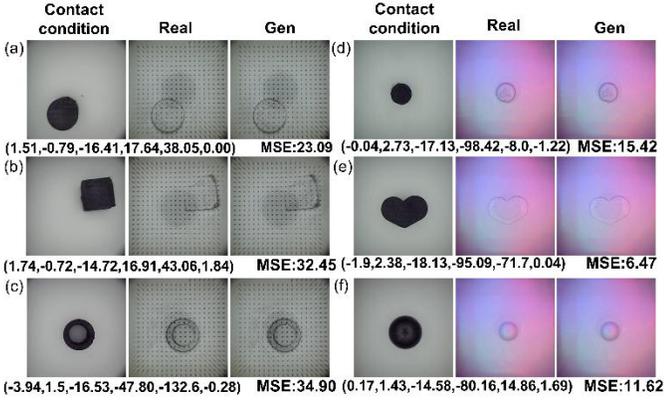

**Fig. 5.** Tactile image generation under different contact conditions. (a) 8mm circle (b) square (c) ring (d) 4mm circle (e) heart (f) sphere.

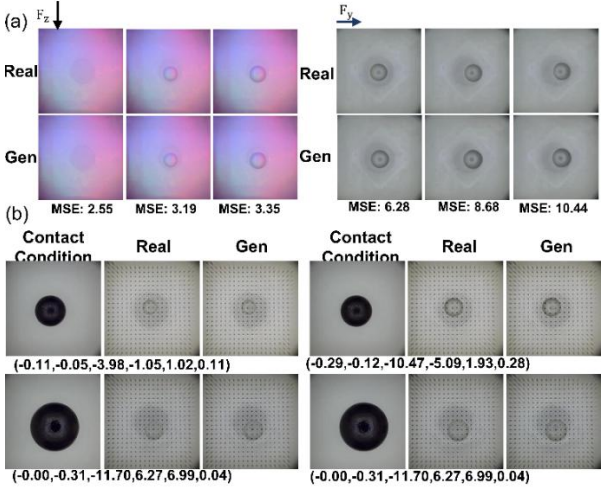

**Fig. 6.** (a) Generated images under different normal and tangential forces (b) Generated images of spheres with different sizes.

MAE reflects the overall brightness deviation, while MSE emphasizes the impact of larger errors, with lower values indicating higher similarity. SSIM primarily assesses the structural similarity of images, considering a combination of luminance, contrast, and structure. PSNR measures the peak signal-to-noise ratio of the image, with higher values indicating better image quality. For different types of vision-based tactile sensors, the generation performance of vision-based tactile images without markers under RGB lighting is the best, with an MSE of 21.00. In contrast, the performance for vision-based tactile images with markers under RGB lighting slightly decreases, with the MSE increasing to 33.54, indicating that the presence of markers affects the complexity of generating simulated tactile images. Under white light, the MSE for vision-based tactile images with markers significantly increases (67.09), and the image quality noticeably deteriorates. This may be due to the relatively flat spectrum of white light, which causes pixel values to be similar and lacks sufficient color contrast, thereby impacting the expression of image details. For the best-performing tactile images without markers under RGB lighting, we compare our results with FOTS [19], which based on lighting model and mechanical model achieving a reduction in MSE by approximately 60.58%. Fig. 5 shows that for objects with different shapes and postures, the model can derive the shape and posture of the object from the provided contact conditions, adjust the pixel distribution, and reconstruct the corresponding vision-based tactile images from the real object images.

**Distinction for different forces:** To investigate the impact of contact force on the generation of tactile images, this section analyzes the model's performance and physical consistency using spheres of different sizes under various force conditions. Fig. 6(a) presents the tactile images of a sphere with a diameter of 10 mm under varying forces. As the normal force increases, the area of the deformation region also expands. Furthermore, the distribution of the shadow in the circular shadow region varies with different tangential forces. Fig. 6(b) depicts spheres with diameters of 10 mm and 15 mm. Since spheres are three-dimensional objects, spheres of different sizes may produce similar shadow areas under certain force conditions. However, the differences in deformation area caused by varying sphere sizes highlight the coupling effect between contact force and geometric dimensions. These experimental results demonstrate that the model can accurately capture differences in contact conditions and generate tactile images consistent with actual contact processes.

**Motion analysis of markers:** In vision-based tactile sensors with markers, the position changes of the markers can reflect variations in contact force and conditions. To evaluate the generation performance of vision-based tactile sensor images with marked points, we introduced the average error metric. This metric aims to provide a clearer understanding of the model's performance through quantitative analysis and assess its ability to differentiate between various contact force conditions. Compared to overall image similarity evaluations, the marker position error offers validation of the method's performance from the perspective of local features. Specifically, this metric calculates the average Euclidean distance $d$ between the positions of markers in the generated and real images, measuring the degree of marker displacement,

$$d = \sum_{i=1,j=1}^{n} \sqrt{\left(x_i^{real} - x_j^{gen}\right)^2 - \left(y_i^{real} - y_j^{gen}\right)^2} \quad (3)$$

where $x_i^{real}$ and $y_i^{real}$ are the coordinates of the centroid position of the $i^{th}$ marked point in the real image, $x_j^{gen}$ and $y_j^{gen}$ are the coordinates of the centroid position of the corresponding marked point in the generated image, with n representing the number of marked points. We detected the marked points in the image using adaptive thresholding and contour detection, selecting $18 \times 18$ marked points in the central area, resulting in a total of 324 marked points. Fig. 7 displays the optical flow images of the generated images alongside those of the real sensor images, confirming that the displacement trends of the marked points in the generated images are consistent with those in the real images. Table 2 summarizes the performance results of different models in generating tactile images with markers. For RGB lighting, the proposed method achieves an average marker position error of 91 pixels (approximately 0.28 pixels per marker), which is lower than other methods. Compared to method in [20], our approach achieves a 38.1% reduction in marker displacement error. This indicates that the model can more accurately restore the movement and deformation of the markers. To further analyze and visualize the generation results, optical flow maps

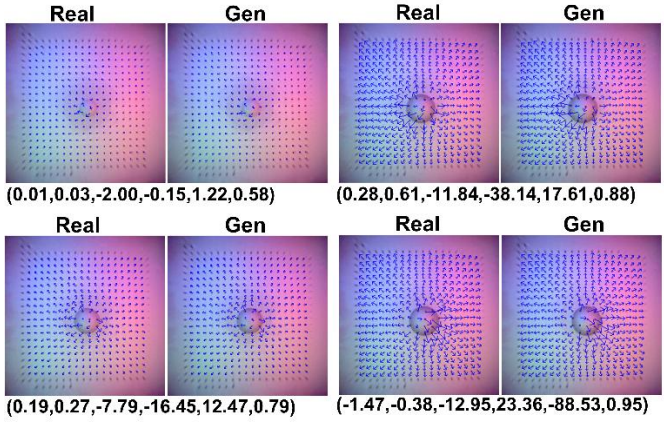

**Fig. 7.** Optical flow of marked points in tactile images under different forces

TABLE II

THE COMPARISON OF THE MARKED POINT DISPLACEMENT ERROR

|  | d(px) |
|---|---|
| Taxim [18] | 205 |
| [20]'s Method | 147 |
| **Ours** | **91** |

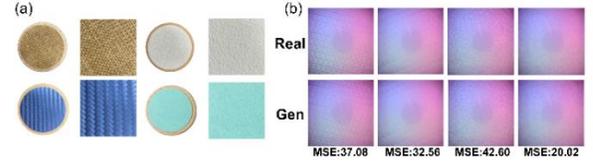

**Fig.8.** (a) Tactile board textures (b) Real tactile images and model-generated images

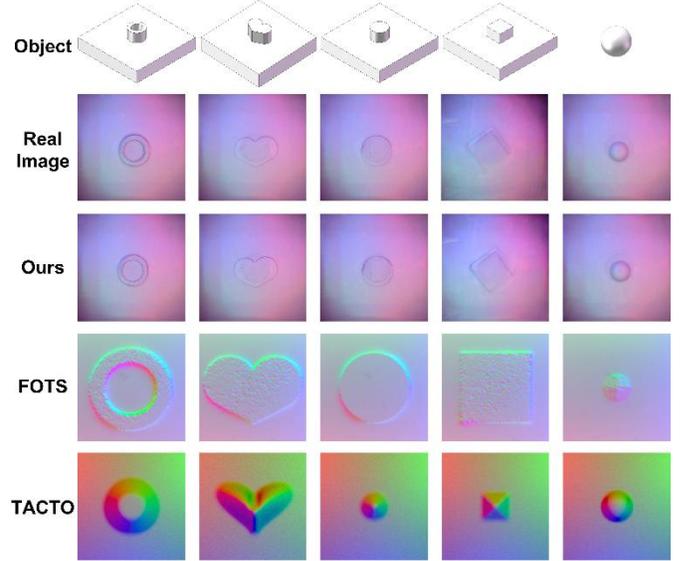

**Fig. 9.** Comparison of visual effects of generated images with those of other works

of the generated marker images were created. Fig. 7 illustrates the generated tactile images and optical flow maps of a sphere under different applied forces. By observing the changes in the optical flow map under varying directions and magnitudes of force, the effectiveness of the proposed method in distinguishing force direction and magnitude in the generated tactile images is more intuitively validated.

*C. Generation of texture features*

In this work, we generated tactile textures for tactile boards used in Montessori children's tactile perception training to demonstrate the model's effectiveness in generating details of tactile images. We restored four different textures from the tactile board (Fig. 8a). The textures of the tactile board can be categorized into regular woven textures and irregular material textures. To avoid the marked points obscuring the textures, we used a three-color light structure of vision-based tactile sensors without marked points to showcase the results of texture feature generation (Fig. 8b).

We evaluated the model's performance by comparing generated images with real images using four metrics: MAE, MSE, SSIM, and PSNR, providing a comprehensive assessment at the pixel level. Additionally, to further analyze the model's performance in generating visual images with markers, we introduced the marker position error metric on top of the image similarity evaluation. This metric validates the model's generation accuracy and consistency from the perspective of local features. These evaluation results demonstrate the effectiveness of the proposed contact condition-guided diffusion model in the task of generating vision-tactile sensor images. Specifically, the model effectively captures changes in contact conditions and generates tactile images that are more consistent with real-world physical scenarios.

Based on the above evaluation, we further compared the generated results of our method with the images generated by the TACTO[28] and Taxim[18] methods in terms of visual effects (Fig. 9). The comparison results show that, under multi-color lighting conditions, the vision-based tactile images generated by our method exhibit more natural shadow distribution in the varying areas and better edge contrast, closely resembling real-world physical effects. In the tactile perception board texture generation task, the images generated by our model based on real data show superior detail performance, particularly demonstrating significant advantages in reproducing complex textures and fine features.

V. CONCLUSION

We propose a condition-guided diffusion method aimed at reducing the need for complex mechanical and optical setups traditionally required for tactile sensors. This method relies on real data to bridge the gap between simulation and reality. Using a contact condition-guided diffusion model, the input consists of object contact images and applied forces, while the output is the corresponding vision-based tactile sensor image, without the need for constructing physical mechanical and optical fields for the sensor's elastomer and emission layer. We assess the quality of the generated images using image similarity metrics and the generation error metrics of marked points. The results show that our method outperforms existing methods in generating vision-based tactile sensor images across various structures. Additionally, the model demonstrates strong performance in generating texture features for tactile perception boards. In summary, our method significantly enhances the quality and detail of generated tactile images, with the potential to advance vision-based tactile sensor simulation technology towards higher accuracy and broader applicability. In the future, this approach is expected to be applied to more complex Sim2Real tasks, such as robotic grasping, tactile feedback in virtual reality, and precise tactile perception in medical devices.